\documentclass[runningheads,a4paper]{llncs}

\usepackage{amssymb,amstext,amsbsy}
\usepackage{graphicx}
\usepackage{subfigure}

\usepackage{algorithm,algorithmic}
\usepackage{multirow}

\begin{document}

\mainmatter  

\title{Multiclass Semi-Supervised Learning on Graphs \\
using Ginzburg-Landau Functional Minimization}

\titlerunning{Multiclass SSL on Graphs using Ginzburg-Landau Functional Minimization}
\toctitle{Multiclass SSL on Graphs using Ginzburg-Landau Functional Minimization}

%
%
\author{Cristina Garcia-Cardona\inst{1}%
\and Arjuna Flenner\inst{2} \and Allon G. Percus\inst{1}}
\authorrunning{Garcia-Cardona et al.}

\tocauthor{Cristina Garcia-Cardona, Arjuna Flenner, and Allon G. Percus}

\institute{Claremont Graduate University,
Institute of Mathematical Sciences, \\ Claremont, CA 91711, USA,\\
cristina.cgarcia@gmail.com, allon.percus@cgu.edu
\and
Naval Air Warfare Center, Physics and Computational Sciences, \\ China Lake, CA 93555, USA}

%
%

\maketitle

\begin{abstract}
We present a graph-based variational algorithm for classification of high-dimensional data, generalizing the binary diffuse interface model to the case of multiple classes.  Motivated by total variation techniques, the method involves minimizing an energy functional made up of three terms.  The first two terms promote a stepwise continuous classification function with sharp transitions between classes, while preserving symmetry among the class labels.  The third term is a data fidelity term, allowing us to incorporate prior information into the model in a semi-supervised framework.  The performance of the algorithm on synthetic data, as well as on the COIL and MNIST benchmark datasets, is competitive with state-of-the-art graph-based multiclass segmentation methods.
\end{abstract}

\newcommand{\Ls}{{\mathbf{L}\boldsymbol{_s}}}

\section{Introduction}
\label{sec:introduction}

Many tasks in pattern recognition and machine learning rely on the ability to quantify local
similarities in data, and to infer meaningful global structure
from such local characteristics~\cite{coifman:lafon:lee}. In the
classification framework, the desired global structure is a descriptive
partition of the data into categories or classes. Many studies
have been devoted to the binary classification problems.  The multiple-class case, where
data are partitioned into more than two clusters, is more
challenging.  One approach is to treat the problem as a series of
binary classification problems~\cite{allwein:schapire:singer}. In this
paper, we develop an alternative method, involving
a multiple-class extension of the diffuse interface
model introduced in~\cite{bertozzi:flenner}. 

The diffuse interface model by Bertozzi and Flenner combines methods for
diffusion on graphs with efficient partial differential equation techniques
to solve binary segmentation problems. As with other methods inspired by
physical phenomena~\cite{bertozzi:esedoglu:gillette,jung:kang:shen,li:kim}, it requires the minimization of an energy expression,
specifically the Ginzburg-Landau (GL) energy functional. The formulation
generalizes the GL functional to the case of functions defined on graphs,
and its minimization is related to the minimization of weighted graph
cuts~\cite{bertozzi:flenner}. In this sense, it parallels other
techniques based on inference on graphs via diffusion operators or
function estimation~\cite{coifman:lafon:lee,chung,zhou:scholkopf,szlam:maggioni:coifman,wang:jebara:chang,buhler:hein,szlam:bresson,hein:setzer}.

Multiclass segmentation methods that cast the problem as a series of
binary classification problems use a number of different strategies:
(i) deal directly with some binary coding or indicator for the
labels~\cite{dietterich:bakiri,wang:jebara:chang}, (ii) build a
hierarchy or combination of classifiers based on the one-vs-all approach
or on class rankings~\cite{hastie:tibshirani,har-peled:roth:zimak} or
(iii) apply a recursive partitioning scheme consisting of successively
subdividing clusters, until the desired number of classes is
reached~\cite{szlam:bresson,hein:setzer}. While there are advantages
to these approaches, such as possible robustness to mislabeled data,
there can be a considerable number of classifiers to compute, and
performance is affected by the number of classes to partition. 

In contrast, we propose an extension of the diffuse interface
model that obtains a simultaneous segmentation into multiple classes. 
The multiclass extension is built by modifying the GL energy functional to
remove the prejudicial effect that the order of the labelings,
given by integer values, has in the smoothing term of the original
binary diffuse interface model. A new term that promotes homogenization
in a multiclass setup is introduced. The expression penalizes data points
that are located close in the graph but are not assigned to the same
class. This penalty is applied {\em independently\/} of how
different the integer values are, representing the class labels.
In this way, the
characteristics of the multiclass classification task are incorporated
directly into the energy functional, with a measure of smoothness
independent of label order, allowing us to obtain high-quality results. Alternative multiclass methods minimize a Kullback-Leibler divergence function~\cite{subramanya:bilmes} or expressions involving the discrete Laplace operator on graphs~\cite{zhou:bousquet:lal,wang:jebara:chang}.

This paper is organized as follows. Section~\ref{sec:model} reviews the
diffuse interface model for binary classification, and describes its
application to
semi-supervised learning. Section~\ref{sec:multiclass} discusses our
proposed multiclass extension and the corresponding computational
algorithm.  Section~\ref{sec:results} presents results obtained with
our method. Finally,
section~\ref{sec:conclusion} draws conclusions and delineates future
work.

\section{Data Segmentation with the Ginzburg-Landau Model}  \label{sec:model}

The diffuse interface model~\cite{bertozzi:flenner} is based on a
continuous approach, using the Ginzburg-Landau (GL) energy functional to
measure the quality of data segmentation. A good segmentation is
characterized by a state with small energy. Let $u(\boldsymbol{x})$ be
a scalar field
defined over a space of arbitrary dimensionality, and representing the
state of the system.  The GL energy is written as the functional
\begin{equation}
\mathrm{GL}(u) = \frac{\epsilon}{2} \int \! | \nabla u |^2 \; d\boldsymbol{x}
+ \frac{1}{\epsilon} \int \! \Phi(u) \; d\boldsymbol{x},  \label{eq:GLf}
\end{equation}
\noindent with $\nabla$ denoting the spatial gradient operator,
$\epsilon > 0$ a real constant value, and $\Phi$ a double well potential with minima at $\pm 1$:
\begin{equation}
	\Phi(u) = \frac{1}{4} \left ( u^2 - 1 \right )^2. \label{eq:2pot}
\end{equation}

Segmentation requires minimizing the GL functional. The norm of the
gradient is a smoothing term that penalizes variations in the
field $u$. The potential term, on the other hand, compels $u$ to adopt
the discrete labels of $+1$ or $-1$, clustering the state of
the system around two classes. Jointly minimizing these two terms pushes
the system domain towards homogeneous regions with values close to the
minima of the double well potential, making the model appropriate for
binary segmentation. 

The smoothing term and potential term are in conflict at the interface
between the two regions, with the first term favoring a gradual
transition, and the second term penalizing deviations from the discrete
labels. A
compromise between these conflicting goals is established via the
constant $\epsilon$. A small value of $\epsilon$ denotes a small length
transition and a sharper interface, while a
large $\epsilon$ weights the gradient norm more, leading to a slower
transition. The result is a diffuse interface between regions, with
sharpness regulated by $\epsilon$. 

It can be shown that in the limit
$\epsilon \to 0$ this function approximates the total variation (TV)
formulation in the sense of functional ($\Gamma$)
convergence~\cite{kohn:sternberg},
producing piecewise constant solutions but with greater computational
efficiency than conventional TV minimization methods.
Thus, the diffuse interface model provides a framework
to compute piecewise constant functions with diffuse transitions,
approaching the ideal of the TV formulation, but with the
advantage that the smooth energy functional is more tractable
numerically and can be minimized by simple numerical methods such
as gradient descent.

The GL energy has been used to approximate the TV norm for image
segmentation~\cite{bertozzi:flenner} and image
inpainting~\cite{bertozzi:esedoglu:gillette,dobrosotskaya:bertozzi_inpainting}.
Furthermore, a calculus on graphs equivalent to TV has been introduced in~\cite{gilboa:osher,szlam:bresson}. 


\subsection*{Application of Diffuse Interface Models to Graphs}

An undirected, weighted neighborhood graph is used to represent the local
relationships in the data set. This is a common technique to segment classes that are not linearly separable.
In the $N$-neighborhood graph model, each vertex $v_i\in V$ of the graph
corresponds to a data point with feature vector $\boldsymbol{x}_i$, while the
weight $w_{ij}$
is a measure of similarity between $v_i$ and $v_j$.
Moreover, it satisfies the symmetry property $w_{ij} = w_{ji}$. The
neighborhood is defined as the set of $N$ closest points in the feature
space. Accordingly, edges exist between each vertex and the vertices of
its $N$-nearest neighbors.  Following the approach of~\cite{bertozzi:flenner},
we calculate weights using the local
scaling of Zelnik-Manor and Perona~\cite{zelnik-manor:perona},
\begin{equation}
	w_{ij} = \exp \left ( - \frac{|| \boldsymbol{x}_i -
\boldsymbol{x}_j ||^2}{\tau(\boldsymbol{x}_i) \;
\tau(\boldsymbol{x}_j)} \right ). \label{eq:local_graph}
\end{equation}
Here, $\tau(\boldsymbol{x}_i) = ||\boldsymbol{x}_i -
\boldsymbol{x}^M_i||$ defines a local value for each $\boldsymbol{x}_i$,
where $\boldsymbol{x}^M_i$ is the position of the
$M$th closest data point to $\boldsymbol{x}_i$, and $M$ is a global
parameter.

It is convenient to express calculations on graphs via the graph
Laplacian matrix, denoted by $\mathbf{L}$.
The procedure we use to build the graph Laplacian is as follows.
\begin{enumerate}
\item Compute the similarity matrix $\mathbf{W}$ with components
$w_{ij}$ defined in (\ref{eq:local_graph}). As the neighborhood
relationship is not symmetric, the resulting matrix $\mathbf{W}$ is also not symmetric. Make it a symmetric matrix by connecting vertices $v_i$ and $v_j$ if $v_i$ is among the $N$-nearest neighbors of $v_j$ or if $v_j$ is among the $N$-nearest neighbors of $v_i$~\cite{luxburg}.
\item Define $\mathbf{D}$ as a diagonal matrix whose $i$th diagonal element represents the degree of the vertex $v_i$, evaluated as
\begin{equation}
  d_i = \sum_{j} w_{ij}.
\end{equation} 
\item Calculate the graph Laplacian: $\mathbf{L} = \mathbf{D} - \mathbf{W}$.
\end{enumerate}
Generally, the graph Laplacian is normalized to guarantee spectral
convergence in the limit of large sample size~\cite{luxburg}.
The symmetric normalized graph Laplacian $\Ls$ is defined
as
\begin{equation}
	\Ls = \mathbf{D}^{-1/2} \; \mathbf{L} \;
\mathbf{D}^{-1/2} = \mathbf{I} - \mathbf{D}^{-1/2} \;
\mathbf{W} \; \mathbf{D}^{-1/2}.
\label{eq:Ls}
\end{equation}

Data segmentation can now be carried out through a graph-based formulation of the GL energy. To implement this task, a fidelity term is added to the functional as initially suggested in~\cite{dobrosotskaya:bertozzi}. This enables the specification of a priori information in the system, for example the known labels of certain points in the data set. This kind of setup is called semi-supervised learning (SSL). The discrete GL energy for SSL on graphs can be written as~\cite{bertozzi:flenner}:
\begin{eqnarray}
\label{eqn:graphLaplacian}
\mathrm{GL}_{\mathrm{SSL}}(\boldsymbol{u}) & = & \frac{\epsilon}{2} \langle
\boldsymbol{u}, \Ls \boldsymbol{u} \rangle + \frac{1}{\epsilon}
\sum_{v_i \in V} \Phi(u(v_i)) + \sum_{v_i \in V} \frac{\mu(v_i)}{2} \; \left ( u(v_i) - \hat{u}(v_i) \right )^2 \\
& = & \frac{\epsilon}{4} \sum_{v_i, v_j \in V} w_{ij} \left ( \frac{u(v_i)}{\sqrt{d_i}} - \frac{u(v_j)}{\sqrt{d_j}} \right )^2 + \frac{1}{\epsilon} \sum_{v_i \in V} \Phi(u(v_i)) + \nonumber \\ 
& &
\sum_{v_i \in V} \frac{\mu(v_i)}{2} \; \left ( u(v_i) - \hat{u}(v_i) \right )^2 .
\end{eqnarray} 
\noindent In the discrete formulation, $\boldsymbol{u}$ is a
vector whose component $u(v_i)$ represents the state of the vertex
$v_i$,
$\epsilon > 0$ is a real constant characterizing the smoothness of the
transition between classes, and $\mu(v_i)$ is a fidelity weight
taking value $\mu > 0$ if the label $\hat{u}(v_i)$ (i.e. class) of the
data point associated with vertex $v_i$ is known beforehand, or
$\mu(v_i)=0$ if
it is not known (semi-supervised).

Minimizing the functional simulates
a diffusion process on the graph. The information of the few labels
known is propagated through the discrete structure by means of the
smoothing term, while the potential term clusters the vertices around
the states $\pm 1$ and the fidelity term enforces the known labels. The energy minimization process itself attempts to reduce the interface regions.
Note that in the absence of the
fidelity term, the process could lead to a trivial
steady-state solution of the diffusion equation, with all data points
assigned the same label.

The final state $u(v_i)$ of each vertex is obtained by thresholding, and the resulting homogeneous regions with labels of $+1$ and $-1$ constitute the two-class data segmentation.

\section{Multiclass Extension}
\label{sec:multiclass}

\noindent The double-well potential in the diffuse interface model for SSL drives the state of the system towards two
definite labels. Multiple-class segmentation requires a more general potential function $\Phi_M(u)$ that allows
clusters around more than two labels. For this purpose, we use the
periodic-well potential suggested by Li and Kim~\cite{li:kim},
\begin{equation}
\Phi_M( u ) = \frac{1}{2} \, \{ u \}^2 \, (\{ u \} - 1)^2, \label{eq:well_ext}
\end{equation}
where $\{ u \}$ denotes the fractional part of $u$,
\begin{equation}
\{ u \} = u - \lfloor u \rfloor, \label{eq:multiphase}
\end{equation}
\noindent and $ \lfloor u \rfloor$ is the largest integer not greater than $u$.

This periodic potential well promotes a multiclass solution, but the
graph Laplacian term in Equation (\ref{eqn:graphLaplacian}) also requires modification for effective calculations due to the fixed ordering of class labels in the multiple class
setting. The graph Laplacian term penalizes large changes in the spatial
distribution of the system state more than smaller gradual
changes. In a multiclass framework, this implies that the penalty for
two spatially contiguous classes with different labels may vary
according to the (arbitrary) ordering of the labels.

\begin{figure}[htb]
\centering
\framebox{\scalebox{0.1}{\includegraphics{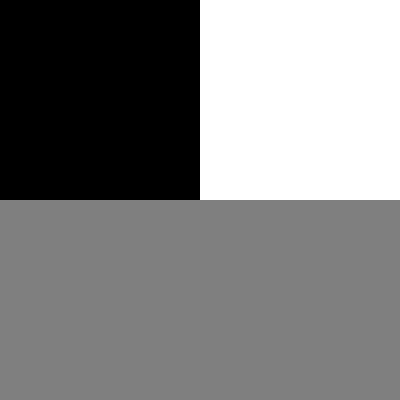}}}
\caption{Three-class segmentation. Black: class 0. Gray: class 1. White: class 2.}
\label{fig:why_multiclass}
\end{figure}

This phenomenon is shown in
Fig.~\ref{fig:why_multiclass}. Suppose that the goal is to segment the
image into three classes: class 0 composed by the black region, class 1
composed by the gray region and class 2 composed by the white region. It
is clear that the horizontal interfaces comprise a jump of size 1
(analogous to a two class segmentation) while the vertical interface
implies a jump of size 2. Accordingly, the smoothing term will assign a
higher cost to the vertical interface, even though from the point of
view of the classification, there is no specific reason for this.  In
this example, the problem cannot be solved with a different label
assignment. There will always be an interface with higher costs than
others independent of the integer values used.

Thus, the multiclass approach breaks the symmetry among classes,
influencing the diffuse interface evolution in an undesirable
manner. Eliminating this inconvenience requires restoring the symmetry,
so that the difference between two classes is always the same,
regardless of their labels. This objective is achieved by
introducing a new class difference measure. 

\subsection{Generalized Difference Function}

The final class labels are determined by thresholding each vertex $u(v_i)$, with the label $y_i$
set to the nearest integer:
\begin{equation}
y_i = \left \lfloor u(v_i) + \frac{1}{2} \right \rfloor.
\end{equation}

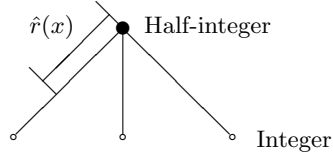
\begin{figure}[h!b]
\begin{center}
\begin{picture}(150,70)(0,0)
\put(60,40){\circle*{5}}
\put(60,40){\line(-1,-1){40}}
\put(60,40){\line(0,-1){40}}
\put(60,40){\line(1,-1){40}}

\put(101,-1){\circle{2}}
\put(60,-1){\circle{2}}
\put(19,-1){\circle{2}}

\put(68,38){\mbox{Half-integer}}
\put(110,-5){\mbox{Integer}}

\put(55,45){\line(-1,-1){25}}
\put(50,50){\line(1,-1){10}}
\put(25,25){\line(1,-1){10}}
\put(25,38){\mbox{$\hat{r}(x)$}}
\end{picture}
\end{center}
\caption{Schematic interpretation of generalized difference:
$\hat{r}(x)$ measures distance to nearest half-integer, and
$\rho$ is a tree distance measure.}
\label{fig:r_hat}
\end{figure}

The boundaries between classes then occur at half-integer values corresponding to the unstable equilibrium states of the potential well.
Define the function $\hat{r}(x)$ to represent the distance to
the nearest half-integer:
\begin{equation}
\hat{r}(x) = \left | \frac{1}{2} - \{ x \} \right |. \label{eq:r_hat}
\end{equation}
A schematic of $\hat{r}(x)$ is depicted in Fig.~\ref{fig:r_hat}. The $\hat{r}(x)$ function is used to define a generalized difference function between
classes that restores symmetry in the energy functional. Define the generalized difference function
$\rho$ as:

\begin{equation}
\rho(u(v_i),u(v_j)) = 
\left \{
\begin{array}{lll}
\hat{r}(u(v_i)) + \hat{r}(u(v_j)) & \ & y_i \neq y_j \\
& & \\
\left|\hat{r}(u(v_i)) - \hat{r}(u(v_j))\right| & & y_i = y_j
\end{array}
\right .
\end{equation}


Thus, if the vertices are in different classes, the difference
$\hat{r}(x)$ between each state's value and the nearest half-integer is added,
whereas if they are in the same
class, these differences are subtracted. The function $\rho(x,y)$ corresponds to
the tree distance (see Fig.~\ref{fig:r_hat}).  Strictly speaking,
$\rho$ is not a metric since it does not satisfy $\rho(x,y) =
0 \Rightarrow  x = y$.  Nevertheless, 
the cost of interfaces between classes becomes the same regardless of class
labeling when this generalized distance function is implemented.

The GL energy functional for SSL, using the
new generalized difference function $\rho$ and the periodic potential,
is expressed as
\begin{eqnarray}
\mathrm{MGL}_{\mathrm{SSL}}(\boldsymbol{u}) & = & \frac{\epsilon}{2}
\sum_{v_i \in V} \sum_{v_j \in V} \frac{w_{ij}}{\sqrt{d_i d_j}} \,
\left[\rho(u(v_i),u(v_j))\,\right]^2 + \nonumber \\
& & \frac{1}{2 \epsilon}\sum_{v_i \in V} \{ u(v_i) \}^2 \, ( \{ u(v_i) \} - 1 )^2 + \nonumber \\
& & \sum_{v_i \in V} \frac{\mu(v_i)}{2} \; \left ( u(v_i) -
\hat{u}(v_i) \right )^2.  \label{eq:multiclass_model}
\end{eqnarray}

Note that the smoothing term in this functional is composed of an operator that is not just a generalization of the normalized symmetric Laplacian $\Ls$. The new smoothing operation, written in terms of the generalized distance function $\rho$, constitutes a non-linear operator that is a symmetrization of a different normalized Laplacian, the random walk Laplacian \mbox{$\mathbf{L}\boldsymbol{_{w}} = \mathbf{D}^{-1}\mathbf{L}=\mathbf{I} - \mathbf{D}^{-1}\mathbf{W}$}~\cite{luxburg}. The reason is as follows. The Laplacian $\mathbf{L}$ satisfies
\begin{displaymath}
     (\mathbf{L}\boldsymbol{u})_{i}= \sum_{j} w_{ij} \left ( u_{i} - u_{j} \right )
\end{displaymath}
and $\mathbf{L}\boldsymbol{_{w}}$ satisfies
\begin{displaymath}
     (\mathbf{L}\boldsymbol{_{w}} \boldsymbol{u})_{i}= \sum_{j} \frac{w_{ij}}{d_i} \left ( u_{i} - u_{j} \right ).
\end{displaymath}
Now replace $w_{ij}/{d_i}$ in the latter expression with the symmetric form ${w_{ij}}/{\sqrt{d_i d_j}}$. 
This is equivalent to constructing a reweighted graph with weights $\widehat{w}_{ij}$ given by:
\begin{displaymath}
	\widehat{w}_{ij} = \frac{w_{ij}}{\sqrt{d_i d_j}}.
\end{displaymath}
The corresponding reweighted Laplacian $\mathbf{\widehat{L}}$ satisfies:
\begin{equation}
     ( \mathbf{\widehat{L}} \boldsymbol{u})_{i} = \sum_{j} \widehat{w}_{ij} \left ( u_{i} - u_{j} \right ) = \sum_{j} \frac{w_{ij}}{\sqrt{d_i d_j}} \left ( u_{i} - u_{j} \right ) ,
\end{equation}
and
\begin{equation}
      \langle \boldsymbol{u},  \mathbf{\widehat{L}} \boldsymbol{u} \rangle = \frac{1}{2} \sum_{i,j} \frac{w_{ij}}{\sqrt{d_i d_j}} \left ( u_{i} - u_{j} \right )^{2} .
\end{equation}

While $\mathbf{\widehat{L}} = \mathbf{\widehat{D}} - \mathbf{\widehat{W}}$ is not a standard normalized Laplacian, it does have the desirable properties of stability and consistency with increasing sample size of the data set, and of satisfying the conditions for $\Gamma$-convergence to TV in the $\epsilon \to 0$ limit~\cite{vangennip:bertozzi}. It also generalizes to the tree distance more easily than does $\Ls$. Replacing the difference $\left ( u_{i} - u_{j} \right )^{2}$ with the generalized difference $\left [ \rho(u_i,u_j) \right ]^2$ 
then gives the new smoothing multiclass term of equation (\ref{eq:multiclass_model}). Empirically, this new term seems to perform well even though the normalization procedure differs from the binary case. 

By implementing the generalized difference function on a tree, the cost of interfaces between classes becomes the same regardless of class labeling. 

\subsection{Computational Algorithm}

The GL energy functional given by (\ref{eq:multiclass_model}) may be
minimized iteratively, using gradient descent:
\begin{equation}
u_i^{n+1} = u_i^{n} - dt \, \left [\frac{\delta
\mathrm{MGL}_{\mathrm{SSL}}}{\delta u_i} \right ],
\end{equation}
where $u_i$ is a shorthand for $u(v_i)$, $dt$ represents the time step and the gradient direction is given by:

\begin{equation}
\frac{\delta \mathrm{MGL}_{\mathrm{SSL}}}{\delta u_i} =  
\epsilon \; \hat{R}(u_i^n) +   \frac{1}{\epsilon} \Phi_M'(u_i^n) + \mu_i \left ( u_i^n - \hat{u}_i \right ) 
\end{equation}

\begin{equation}
\hat{R}(u_i^n) =  \sum_j \frac{w_{ij}}{\sqrt{d_i d_j}} \left [ \hat{r}(u_i^n) \pm \hat{r}(u_j^n) \right ] \hat{r}'(u_i^n) \label{eq:G} 
\end{equation}

\begin{equation}
\Phi_M'(u_i^n) = 2 \; \{ u_i^n \} ^3 - 3 \; \{ u_i^n\} ^2 + \{ u_i^n \}
\end{equation}

The gradient of the generalized difference function $\rho$ is not defined at half integer values.  Hence, we
modify the method using a greedy strategy: after detecting that a
vertex changes class, the new class that minimizes the
smoothing term is selected, and the fractional part of the state
computed by the gradient descent update is preserved. Consequently, the
new state of vertex $i$ is the result of gradient descent,
but if this causes a change in class, then a new state is determined.

\begin{algorithm*}[!b]
\caption{Calculate $\boldsymbol{u}$} \label{algo:iter}
\begin{algorithmic} 
\REQUIRE $\epsilon, dt, N_D, n_{\mathrm{max}}, K$
\ENSURE $\mathrm{out} = \boldsymbol{u^{\mathrm{end}}}$
\FOR{$i = 1  \to N_D$}
\STATE $u_i^{~0} \leftarrow rand((0,K))-\frac{1}{2}. \quad \mathrm{If~} \mu_i > 0, ~ u_{i}^{~0} \leftarrow
\hat{u}_{i}$
\ENDFOR
\FOR{$n = 1 \to n_{\mathrm{max}}$}
\FOR{$i = 1  \to N_D$}
\STATE $u_i^{n+1} \leftarrow u_i^n - dt \left ( \epsilon \: \hat{R}(u_i^n)  +
\frac{1}{\epsilon} \: \Phi_M'(u_i^n) + \mu_i \left ( u_i^n - \hat{u}_i \right ) \right )$
\IF{$\mathrm{Label}(u_i^{n+1}) \neq  \mathrm{Label}(u_i^{n})$}
\STATE $\hat{k}=\arg\min_{\; 0 \leq k < K} \; \sum_{j} \frac{w_{ij}}{\sqrt{d_i d_j}} \,
\left[\rho(k + \{ u_i ^{n+1}\}, u_j^{n+1})\,\right]^2$
\STATE $u_i ^{n+1} \leftarrow \hat{k} + \{ u_i ^{n+1}\}$
\ENDIF
\ENDFOR
\ENDFOR
\end{algorithmic}
\end{algorithm*}

Specifically, let
$k$ represent an integer in the range of the problem, i.e. $ k \in [0,
K-1]$, where $K$ is the number of classes in the problem. Given the
fractional part $\{u\}$ resulting from the gradient descent update, find the integer $k$ that minimizes
$\sum_{j} \frac{w_{ij}}{\sqrt{d_i d_j}} \,
\left[\rho(k + \{ u_i\}, u_j)\,\right]^2$, the smoothing term in the energy functional, and use $k + \{ u_i\}$ as
the new vertex state.
A summary of the procedure is shown in Algorithm~\ref{algo:iter}
with $N_D$ representing the number of points in the data set and $n_{\mathrm{max}}$ denoting the maximum number of iterations.

\section{Results}
\label{sec:results}

The performance of the multiclass diffuse interface model
is evaluated using a number of data sets from the literature, with differing
characteristics.  Data and image segmentation problems are considered on synthetic and real data sets.

\subsection{Synthetic Data}

\subsubsection{Three Moons.} \label{sec:3moons}
A synthetic three-class segmentation problem is constructed following an analogous procedure to the one used in~\cite{buhler:hein} for ``two moon'' binary classification. Three half circles (``three moons'') are generated in $\mathbb{R}^2$. The two top circles have radius $1$ and are centered at $(0, 0)$ and $(3, 0)$. The bottom half circle has radius $1.5$ and is centered at $(1.5, 0.4)$. 1,500 data points (500 from each of these half circles) are sampled and embedded in $\mathbb{R}^{100}$. The embedding is completed by adding Gaussian noise with $\sigma^2= 0.02$ to {\em each\/} of the 100 components for each data point. The dimensionality of the data set, together with the noise, make this a nontrivial problem. 


The symmetric normalized graph Laplacian is computed for a local scaling graph using $N = 10$ nearest neighbors and local scaling based on the $M = 10^{th}$ closest point. The fidelity term is constructed by labeling 25 points per class, 75 points in total, corresponding to only 5\% of the points in the data set.  The multiclass GL method was further refined by geometrically decreasing $\epsilon$ over the course of the minimization process, from $\epsilon_0$ to $\epsilon_f$ by factors of $1-\Delta_\epsilon$ ($n_{\mathrm{max}}$ iterations per value of $\epsilon$), to allow sharper transitions between states as in~\cite{bertozzi:flenner}.  Table~\ref{tab:res_3moons} specifies the parameters used. Average accuracies and computation times are reported over 100 runs. Results for $k$-means and spectral clustering (obtained by applying $k$-means to the first 3 eigenvectors of $\mathbf{L}\boldsymbol{_{s}}$)  are included as reference.

\begin{table}[!b]
\centering
\caption{Three-moons results}
\label{tab:res_3moons}
\renewcommand{\arraystretch}{1.3}
\begin{tabular}{ | p{3.4 cm} | l | c | c |}
\hline
{\bf Method} & {\bf Parameters} & {\bf Correct \% (stddev \%)} & {\bf Time [s]} \\
\hline
\hline
$k$-means & -- & 72.1 (0.35) & 0.66 \\ 
\hline
Spectral clustering & 3 eigenvectors & 80.0 (0.59) & 0.02 \\ 
\hline
\multirow{2}{3.4cm}{Multiclass GL} & ${\mu = 30}$, ${\epsilon = 1}$, ${dt = 0.01}$, & \multirow{2}{*}{95.1 (2.33)} & \multirow{2}{*}{0.89} \\ 
 & ${n_{\mathrm{max}} = 1,000}$ & & \\
\hline
\multirow{3}{3.4cm}{Multiclass GL (adaptive $\epsilon$)} & ${\mu = 30}, {\epsilon_0 = 2}, {\epsilon_f = 0.01},$ & \multirow{3}{*}{96.2 (1.59)} & \multirow{3}{*}{1.61} \\ 
& ${\Delta_{\epsilon} = 0.1}, {dt = 0.01},$ & & \\
& ${n_{\mathrm{max}} = 40}$ & & \\
\hline
\end{tabular}
\end{table}

\begin{figure}[t]
\centerline{
\subfigure{\scalebox{0.2}{\includegraphics{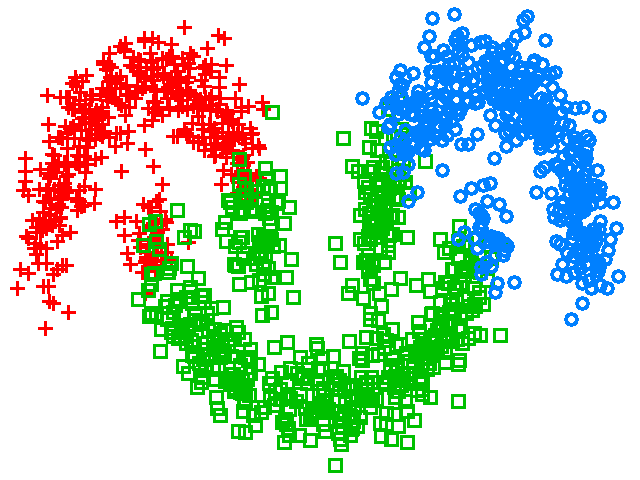}}}
\hfil
\subfigure{\scalebox{0.2}{\includegraphics{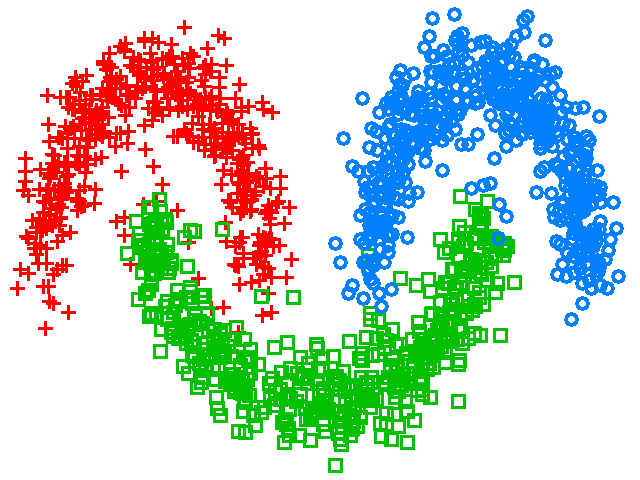}}}
}
\caption[Three-moons segmentation]{Three-moons segmentation. Left: spectral clustering. Right: multiclass GL with adaptive $\epsilon$.}
\label{fig:3moon_res}
\end{figure}

Segmentations obtained for spectral clustering and for multiclass GL with adaptive $\epsilon$ methods are shown in Fig.~\ref{fig:3moon_res}. The figure displays the \emph{best} result obtained over 100 runs, corresponding to accuracies of $81.3\%$ (spectral clustering) and 97.9\% (multiclass GL with adaptive $\epsilon$). The same graph structure is used for the spectral clustering decomposition and the multiclass GL method.

\begin{figure}[t]
\centerline{
\subfigure[100 iterations]{\scalebox{0.2}{\includegraphics{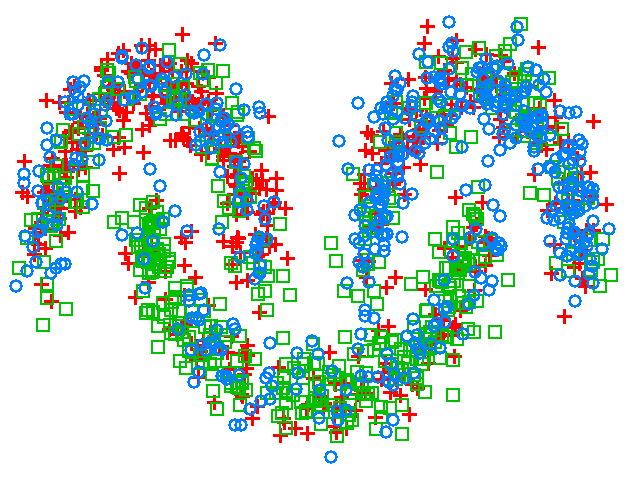}}}
\hfil
\subfigure[300 iterations]{\scalebox{0.2}{\includegraphics{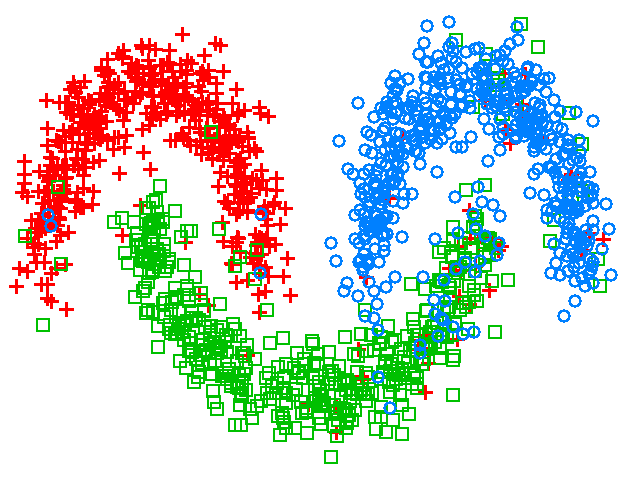}}}
}
\vfil
\centerline{
\subfigure[1,000 iterations]{\scalebox{0.2}{\includegraphics{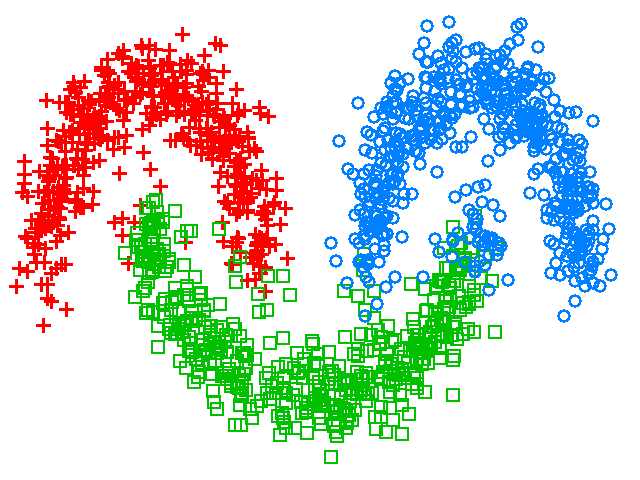}}}
\hfil
\subfigure[Energy evolution]{\scalebox{0.2}{\includegraphics{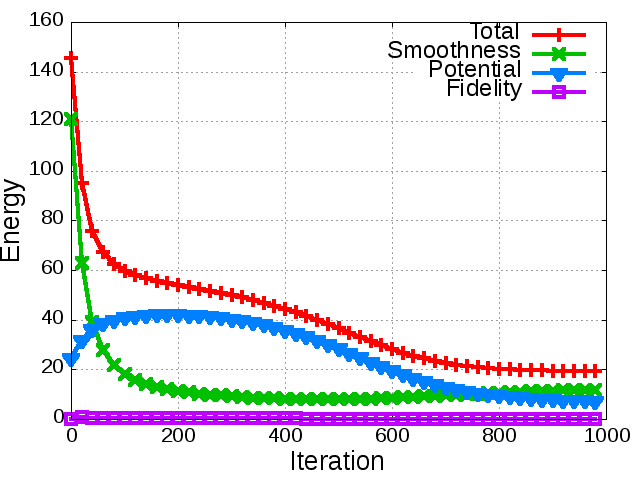}}
\label{fig:3moons_scalar_energy}}
}
\caption{Evolution of label values in three moons, using multiclass GL (fixed $\epsilon$): $\mathbb{R}^2$ projections at 100, 300 and 1,000 iterations, and energy evolution.}
\label{fig:res_3moon_evol}
\end{figure}

For comparison, we note the results from the literature for the simpler two-moon problem (also $\mathbb{R}^{100}$, $\sigma^2= 0.02$ noise). The best results reported include: 94\% for $p$-Laplacian~\cite{buhler:hein}, 95.4\% for ratio-minimization relaxed Cheeger cut~\cite{szlam:bresson}, and 97.7\% for binary GL~\cite{bertozzi:flenner}. While these are not SSL methods, the last of these does involve other prior information in the form of a mass balance constraint.
It can be seen that our procedures produce similarly high-quality results even for the more complex three-class segmentation problem. 

It is instructive to observe the evolution of label values in the multiclass method. Fig.~\ref{fig:res_3moon_evol} displays $\mathbb{R}^2$ projections of the results of multiclass GL (with fixed $\epsilon$), at 100, 300 and 1,000 iterations. The system starts from a random configuration.  Notice that after 100 iterations, the structure is still fairly inhomogeneous, but small uniform regions begin to form. These correspond to islands around fidelity points and become seeds for further homogenization. The system progresses fast, and by 300 iterations the configuration is close to the final result: some points are still incorrectly labeled, mostly on the boundaries, but the classes form nearly uniform clusters. By 1,000 iterations the procedure converges to a steady state and a high-quality multiclass segmentation (95\% accuracy) is obtained.

In addition, the energy evolution for one typical run is shown in Fig.~\ref{fig:3moons_scalar_energy} for the case with fixed $\epsilon$. The figure includes plots of the total energy (red) as well as the partial contributions of each of the three terms, namely smoothing (green), potential (blue) and fidelity (purple). Observe that at the initial iterations, the principal contribution to the energy comes from the smoothing term, but it has a fast decay due to the homogenization taking place. At the same time, the potential term increases, as $\rho$ pushes the label values toward half-integers. Eventually, the minimization process is driven by the potential term, while small local adjustments are made. The fidelity term is satisfied quickly and has almost negligible influence after the first few iterations.  This picture of the ``typical'' energy evolution can serve as a useful guide in evaluating the performance of the method when no ground truth is available. 

\subsubsection{Swiss Roll.} \label{sec:swissroll}

\begin{table}[!b]
\centering
\caption{Swiss roll results}
\label{tab:res_swiss_roll}
\renewcommand{\arraystretch}{1.3}
\begin{tabular}{ | l | l | c | c |}
\hline
{\bf Method} & {\bf Parameters} & {\bf Correct \% (stddev \%)} & {\bf Time [s]} \\
\hline
\hline
$k$-means & -- & 37.9 (0.91) & 0.05 \\ 
\hline
Spectral Clustering & 4 eigenvectors & 49.7 (0.96) & 0.05 \\ 
\hline
\multirow{2}{3.5cm}{Multiclass GL} & ${\mu = 50}$, ${\epsilon = 1}$, ${dt = 0.01}$ & \multirow{2}{*}{91.0 (2.72)} & \multirow{2}{*}{0.75} \\ 
& ${n_{\mathrm{max}} = 1,000}$ & & \\
\hline
\end{tabular}
\end{table}

A synthetic four-class segmentation problem is constructed using the Swiss roll mapping, following the procedure in~\cite{surendran}. The data are created in $\mathbb{R}^2$ by randomly sampling from a Gaussian mixture model of four components with means at $(7.5,7.5)$, $(7.5,12.5)$, $(12.5,7.5)$ and $(12.5,12.5)$, and all covariances given by the $2 \times 2$ identity matrix. 1,600 points are sampled (400 from each of the Gaussians).
The data are then converted from 2 to 3 dimensions, with the following Swiss roll mapping: $(x,y) \rightarrow (x \cos(x), y, x \sin(x))$. 


As before, we construct the weight matrix for a local scaling graph, with $N=10$ and scaling based on the $M=10^{th}$ closest neighbor. The fidelity set is formed by labeling $5$\% of the points selected randomly.

\begin{figure}[!t]
\vspace{-0.6cm}
\centerline{
\subfigure[Spectral clustering]{\scalebox{0.25}{\includegraphics{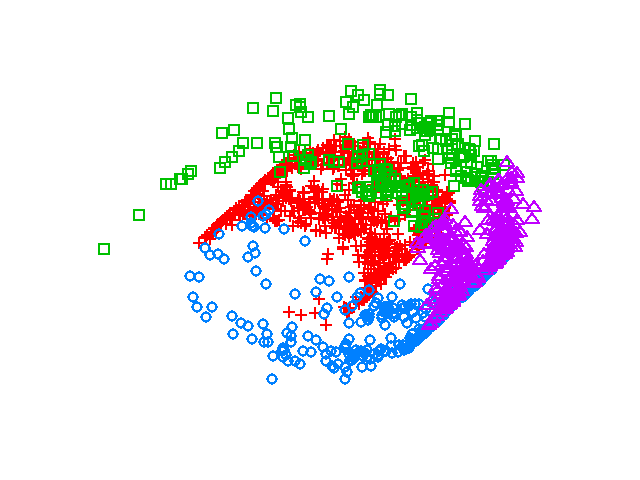}}}
\hfil
\subfigure[Multiclass GL]{\scalebox{0.25}{\includegraphics{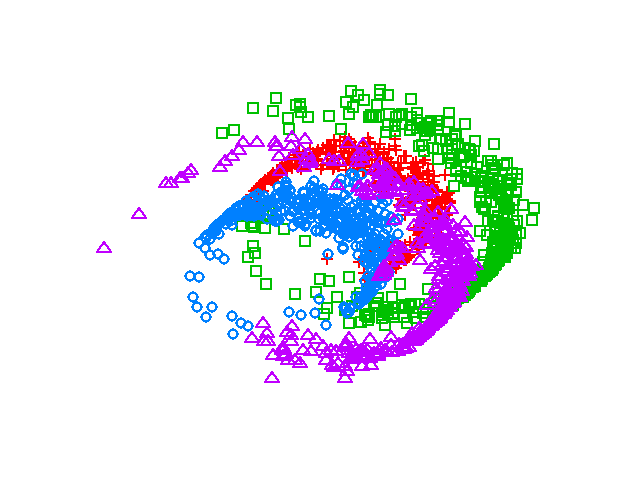}}}
}
\caption{Swiss roll results.}
\label{fig:swissroll_results}
\end{figure}

Table~\ref{tab:res_swiss_roll} gives a description of the parameters used, as well as average results over 100 runs for $k$-means, spectral clustering and multiclass GL. The \emph{best} results achieved over these 100 runs are shown in Fig.~\ref{fig:swissroll_results}. These correspond to accuracies of 
$50.1\%$ (spectral clustering) and $96.4\%$ (multiclass GL). Notice that spectral clustering produces results composed of compact classes, but with a configuration that does not follow the manifold structure. In contrast, the multiclass GL method is capable of segmenting the manifold structure correctly, achieving higher accuracies. 

\subsection{Image Segmentation}

\begin{figure}
\centerline{
\subfigure[Original]{\scalebox{0.3}{\includegraphics{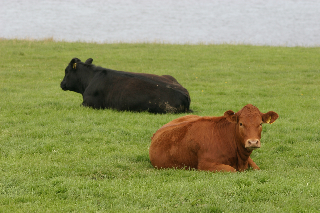}}
\label{fig:cows_original}}
\hfil
\subfigure[Sampled]{\scalebox{0.3}{\includegraphics{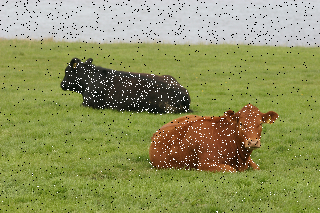}}
\label{fig:cows_sampled}}
\hfil
\subfigure[Black cow]{\scalebox{0.3}{\includegraphics{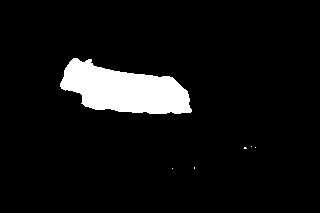}}
\label{fig:cows_0}}
}
\vfil
\centerline{
\subfigure[Red cow]{\scalebox{0.3}{\includegraphics{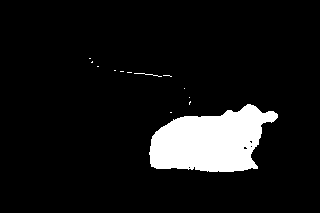}}}
\hfil
\subfigure[Grass]{\scalebox{0.3}{\includegraphics{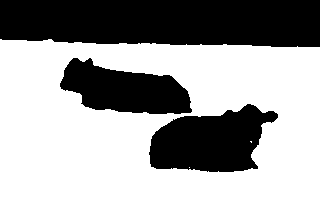}}}
\hfil
\subfigure[Sky]{\scalebox{0.3}{\includegraphics{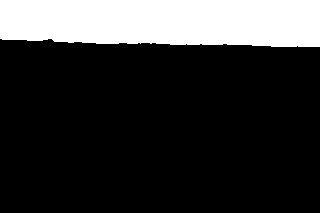}}
\label{fig:cows_3}}
}
\caption{Color (multi-channel) image. Original image, sampled fidelity and results.}
\label{fig:cows}
\end{figure}

We apply our algorithm to the color image of cows shown in Fig.~\ref{fig:cows_original}.  This is a $213 \times 320$ color image, to be divided into four classes: sky, grass, black cow and red cow. To construct the weight matrix, we use feature vectors defined as the set of intensity values in the neighborhood of a pixel. The neighborhood is a patch of size $5 \times 5$. Red, green and blue channels are appended, resulting in a feature vector of dimension 75. A local scaling graph with $N = 30$ and $M=30$ is constructed.  For the fidelity term, 2.6\% of labeled pixels are used (Fig.~\ref{fig:cows_sampled}).

The multiclass GL method used the following parameters: $\mu=30$, $\epsilon=1$, $dt=0.01$ and $n_{\mathrm{max}}=800$. The average time for segmentation using different fidelity sets was $19.9$ s. Results are depicted in Figs.~\ref{fig:cows_0}-\ref{fig:cows_3}. Each class image shows in white the pixels identified as belonging to the class, and in black the pixels of the other classes. It can be seen that all the classes are clearly segmented.  The few mistakes made are in identifying some borders of the black cow as part of the red cow, and vice-versa.

\pagebreak

\subsection{Benchmark Sets}
\subsubsection{COIL-100.} \label{sec:coil}
The Columbia object image library (COIL-100) is a set of 7,200 color images of 100 different objects taken from different angles (in steps of 5 degrees) at a resolution of $128 \times 128$ pixels~\cite{coil}. This image database has been preprocessed and made available by~\cite{chapelle:scholkopf:zien} as a benchmark for SSL algorithms. In summary, the red channel of each image is downsampled to $16 \times 16$ pixels by averaging over blocks of $8 \times 8$ pixels. Then $24$ of the objects are randomly selected and partitioned into six arbitrary classes: $38$ images are discarded from each class, leaving $250$ per class or 1,500 images in all. The downsampled $16 \times 16$ images are further processed to hide the image structure by rescaling, adding noise and masking 15 of the 256 components. The result is a data set of 1,500 data points, of dimension 241.

We build a local scaling graph, with $N = 4$ nearest neighbors and scaling based on the $M=4^{th}$ closest neighbor. The fidelity term is constructed by labeling $10$\% of the points, selected at random. The multiclass GL method used the following parameters: $\mu=100$, $\epsilon=4$, $dt=0.02$ and $n_{\mathrm{max}}=$ 1,000. An average accuracy of 93.2\%, with standard deviation of 1.27\%, is obtained over 100 runs, with an average time for segmentation of $0.29$s.

For comparison, we note the results reported in~\cite{subramanya:bilmes}: 83.5\% ($k$-nearest neighbors), 87.8\% (LapRLS), 89.9\% (sGT), 90.9\% (SQ-Loss-I) and 91.1\% (MP). All these are SSL methods (with the exception of $k$-nearest neighbors which is supervised), using 10\% fidelity just as we do. As can be seen, our results are of greater accuracy.

\subsubsection{MNIST Data.}
The MNIST data set~\cite{lecun:cortes} is composed of 70,000  $28 \times 28$ images of handwritten digits $0$ through $9$. The task is to classify each of the images into the corresponding digit.  Hence, this is a $10$-class segmentation problem.

The weight matrix constructed corresponds to a local scaling graph with $N = 8$ nearest neighbors and scaling based on the $M=8^{th}$ closest neighbor. We perform no preprocessing, so the graph directly uses the $28 \times 28$ images. This yields a data set of 70,000 points of dimension 784. For the fidelity term, 250 images per class (2,500 images, corresponding to $3.6 \%$ of the data) are chosen randomly. The multiclass GL method used the following parameters: $\mu=50$, $\epsilon=1$, $dt=0.01$ and $n_{\mathrm{max}}=$ 1,500. An average accuracy of 96.9\%, with standard deviation of 0.04\%, is obtained over 50 runs. The average time for segmentation using different fidelity sets was $60.89$ s.

Comparative results from other methods reported in the literature include: 87.1\% (p-Laplacian~\cite{buhler:hein}),  87.64\% (multicut normalized 1-cut~\cite{hein:setzer}), 88.2\% (Cheeger cuts~\cite{szlam:bresson}), 92.6\% (transductive classification~\cite{szlam:maggioni:coifman}).  As with the three-moon problem, some of these are based on unsupervised methods but incorporate enough prior information that they can fairly be compared with SSL methods.  Comparative results from {\em supervised\/} methods are: 88\% (linear classifiers~\cite{lecun,lecun:cortes}), 92.3-98.74\% (boosted stumps~\cite{lecun:cortes}), 95.0-97.17\% ($k$-nearest neighbors~\cite{lecun,lecun:cortes}),  95.3-99.65\% (neural/convolutional nets~\cite{lecun,lecun:cortes}), 96.4-96.7\% (nonlinear classifiers~\cite{lecun,lecun:cortes}), 98.75-98.82\% (deep belief nets~\cite{hinton:osindero:teh}) and 98.6-99.32\% (SVM~\cite{lecun}).
Note that all of these take 60,000 of the digits as a training set and 10,000 digits as a testing set~\cite{lecun:cortes}, in comparison to our approach where we take only $3.6\%$ of the points for the fidelity term.  Our SSL method is nevertheless competitive with these supervised methods. Moreover, we perform no preprocessing or initial feature extraction on the image data, unlike most of the other methods we compare with (we have excluded from the comparison, however, methods that explicitly deskew the image). While there is a computational price to be paid in forming the graph when data points
use all 784 pixels as features, this is a simple one-time operation.

\section{Conclusions}
\label{sec:conclusion}

\noindent We have proposed a new multiclass segmentation procedure, based on the
diffuse interface model. The method obtains segmentations of
several classes simultaneously without using one-vs-all or alternative
sequences of binary segmentations required by other multiclass methods.
The local scaling method of Zelnik-Manor and Perona, used to construct
the graph, constitutes a useful representation of the characteristics of
the data set and is adequate to deal with high-dimensional data. 

Our modified diffusion method, represented by the non-linear smoothing
term introduced in the Ginzburg-Landau functional, exploits the
structure of the multiclass model and is not affected by the ordering of
class labels.  It efficiently propagates class information that is known
beforehand, as evidenced by the small proportion of fidelity points (2\%
-- 10\% of dataset) needed to perform accurate segmentations.  Moreover, the
method is robust to initial conditions.  As long as the initialization
represents all classes uniformly, different initial random configurations
produce very similar results.  The main limitation of the method appears
to be that fidelity points must be representative of class distribution.
As long as this holds, such as in the examples discussed, the
long-time behavior of the solution relies less on choosing the ``right''
initial conditions than do other learning techniques on graphs.

State-of-the-art results with small classification errors were obtained
for all classification tasks. Furthermore, the results do not depend on
the particular class label assignments. Future work includes
investigating the diffuse interface parameter $\epsilon$.  We conjecture
that the proposed functional converges (in the $\Gamma$-convergence
sense) to a total variational type functional on graphs as $\epsilon$ approaches zero, but the exact nature of the limiting functional is unknown. 

\subsubsection*{Acknowledgements.}
This research has been supported by the Air Force Office of Scientific Research MURI grant FA9550-10-1-0569 and by ONR grant  N0001411AF00002.

\bibliographystyle{splncs_srt}
\bibliography{ref_graph}

\begin{thebibliography}{10}

\bibitem{allwein:schapire:singer}
Allwein, E.L., Schapire, R.E., Singer, Y.:
\newblock Reducing multiclass to binary: A unifying approach for margin
  classifiers.
\newblock Journal of Machine Learning Research \textbf{1} (2000)  113--141

\bibitem{vangennip:bertozzi}
Bertozzi, A., van Gennip, Y.:
\newblock Gamma-convergence of graph {G}inzburg-{L}andau functionals.
\newblock Advanced in Differential Equations \textbf{17}(11--12) (2012)
  1115--1180

\bibitem{bertozzi:esedoglu:gillette}
Bertozzi, A., Esedo{\=g}lu, S., Gillette, A.:
\newblock Inpainting of binary images using the {C}ahn-{H}illiard equation.
\newblock IEEE Transactions on Image Processing \textbf{16}(1) (2007)  285--291

\bibitem{bertozzi:flenner}
Bertozzi, A.L., Flenner, A.:
\newblock Diffuse interface models on graphs for classification of high
  dimensional data.
\newblock Multiscale Modeling and Simulation \textbf{10}(3) (2012)  1090--1118

\bibitem{buhler:hein}
B\"uhler, T., Hein, M.:
\newblock Spectral clustering based on the graph $p$-{L}aplacian.
\newblock In Bottou, L., Littman, M., eds.: Proceedings of the 26th
  International Conference on Machine Learning.
\newblock Omnipress, Montreal, Canada (2009)  81--88

\bibitem{chapelle:scholkopf:zien}
Chapelle, O., Sch{\"o}lkopf, B., Zien, A., eds.:
\newblock Semi-Supervised Learning.
\newblock MIT Press, Cambridge, MA (2006)

\bibitem{chung}
Chung, F.R.K.:
\newblock Spectral graph theory.
\newblock In: Regional Conference Series in Mathematics. Volume~92.
\newblock Conference Board of the Mathematical Sciences (CBMS), Washington, DC
  (1997)

\bibitem{coifman:lafon:lee}
Coifman, R.R., Lafon, S., Lee, A.B., Maggioni, M., Nadler, B., Warner, F.,
  Zucker, S.W.:
\newblock Geometric diffusions as a tool for harmonic analysis and structure
  definition of data: Diffusion maps.
\newblock Proceedings of the National Academy of Sciences \textbf{102}(21)
  (2005)  7426--7431

\bibitem{dietterich:bakiri}
Dietterich, T.G., Bakiri, G.:
\newblock Solving multiclass learning problems via error-correcting output
  codes.
\newblock Journal of Artificial Intelligence Research \textbf{2}(1) (1995)
  263--286

\bibitem{dobrosotskaya:bertozzi_inpainting}
Dobrosotskaya, J.A., Bertozzi, A.L.:
\newblock A wavelet-{L}aplace variational technique for image deconvolution and
  inpainting.
\newblock IEEE Trans. Image Process. \textbf{17}(5) (2008)  657--663

\bibitem{dobrosotskaya:bertozzi}
Dobrosotskaya, J.A., Bertozzi, A.L.:
\newblock Wavelet analogue of the {G}inzburg-{L}andau energy and its
  gamma-convergence.
\newblock Interfaces and Free Boundaries \textbf{12}(2) (2010)  497--525

\bibitem{gilboa:osher}
Gilboa, G., Osher, S.:
\newblock Nonlocal operators with applications to image processing.
\newblock Multiscale Modeling and Simulation \textbf{7}(3) (2008)  1005--1028

\bibitem{har-peled:roth:zimak}
Har-Peled, S., Roth, D., Zimak, D.:
\newblock Constraint classification for multiclass classification and ranking.
\newblock In S.~Becker, S.T., Obermayer, K., eds.: Advances in Neural
  Information Processing Systems 15.
\newblock MIT Press, Cambridge, MA (2003)  785--792

\bibitem{hastie:tibshirani}
Hastie, T., Tibshirani, R.:
\newblock Classification by pairwise coupling.
\newblock In: Advances in Neural Information Processing Systems 10.
\newblock MIT Press, Cambridge, MA (1998)

\bibitem{hein:setzer}
Hein, M., Setzer, S.:
\newblock Beyond spectral clustering - tight relaxations of balanced graph
  cuts.
\newblock In Shawe-Taylor, J., Zemel, R., Bartlett, P., Pereira, F.,
  Weinberger, K., eds.: Advances in Neural Information Processing Systems 24.
\newblock (2011)  2366--2374

\bibitem{hinton:osindero:teh}
Hinton, G.E., Osindero, S., Teh, Y.W.:
\newblock A fast learning algorithm for deep belief nets.
\newblock Neural Computation \textbf{18} (2006)  1527--1554

\bibitem{jung:kang:shen}
Jung, Y.M., Kang, S.H., Shen, J.:
\newblock Multiphase image segmentation via {M}odica-{M}ortola phase
  transition.
\newblock SIAM J. Appl. Math \textbf{67}(5) (2007)  1213--1232

\bibitem{kohn:sternberg}
Kohn, R.V., Sternberg, P.:
\newblock Local minimizers and singular perturbations.
\newblock Proc. Roy. Soc. Edinburgh Sect. A \textbf{111}(1-2) (1989)  69--84

\bibitem{lecun}
LeCun, Y., Bottou, L., Bengio, Y., Haffner, P.:
\newblock Gradient-based learning applied to document recognition.
\newblock Proceedings of the IEEE \textbf{86}(11) (1998)  2278--2324

\bibitem{lecun:cortes}
LeCun, Y., Cortes, C.:
\newblock The {MNIST} database of handwritten digits.
\newblock http://yann.lecun.com/exdb/mnist/

\bibitem{li:kim}
Li, Y., Kim, J.:
\newblock Multiphase image segmentation using a phase-field model.
\newblock Computers and Mathematics with Applications \textbf{62} (2011)
  737--745

\bibitem{coil}
Nene, S., Nayar, S., Murase, H.:
\newblock {C}olumbia {O}bject {I}mage {L}ibrary ({COIL}-100).
\newblock Technical Report CUCS-006-96 (1996)

\bibitem{subramanya:bilmes}
Subramanya, A., Bilmes, J.:
\newblock Semi-supervised learning with measure propagation.
\newblock Journal of Machine Learning Research \textbf{12} (2011)  3311--3370

\bibitem{surendran}
Surendran, D.:
\newblock Swiss roll dataset.
\newblock http://people.cs.uchicago.edu/\~{}dinoj/manifold/ swissroll.html
  (2004)

\bibitem{szlam:bresson}
Szlam, A., Bresson, X.:
\newblock Total variation and cheeger cuts.
\newblock In F\"urnkranz, J., Joachims, T., eds.: Proceedings of the 27th
  International Conference on Machine Learning.
\newblock Omnipress, Haifa, Israel (2010)  1039--1046

\bibitem{szlam:maggioni:coifman}
Szlam, A.D., Maggioni, M., Coifman, R.R.:
\newblock Regularization on graphs with function-adapted diffusion processes.
\newblock Journal of Machine Learning Research \textbf{9} (2008)  1711--1739

\bibitem{luxburg}
von Luxburg, U.:
\newblock A tutorial on spectral clustering.
\newblock Technical Report TR-149, Max Planck Institute for Biological
  Cybernetics (2006)

\bibitem{wang:jebara:chang}
Wang, J., Jebara, T., Chang, S.F.:
\newblock Graph transduction via alternating minimization.
\newblock Proceedings of the 25th International Conference on Machine Learning
  (2008)

\bibitem{zelnik-manor:perona}
{Zelnik-Manor}, L., {Perona}, P.:
\newblock Self-tuning spectral clustering.
\newblock In Saul, L.K., Weiss, Y., Bottou, L., eds.: Advances in Neural
  Information Processing Systems 17.
\newblock MIT Press, Cambridge, MA (2005)

\bibitem{zhou:bousquet:lal}
Zhou, D., Bousquet, O., Lal, T.N., Weston, J., Sch\"olkopf, B.:
\newblock Learning with local and global consistency.
\newblock In Thrun, S., Saul, L.K., Sch\"olkopf, B., eds.: Advances in Neural
  Information Processing Systems 16.
\newblock MIT Press, Cambridge, MA (2004)  321--328

\bibitem{zhou:scholkopf}
Zhou, D., Sch\"olkopf, B.:
\newblock A regularization framework for learning from graph data.
\newblock In: Workshop on Statistical Relational Learning.
\newblock International Conference on Machine Learning, Banff, Canada (2004)

\end{thebibliography}

\end{document}